\pdfoutput=1

\documentclass[11pt]{article}

\usepackage{authblk}
\usepackage[hyperref]{acl2021}
\def\aclpaperid{776}

\usepackage{times}
\usepackage{latexsym}
\usepackage{float}
\usepackage[T1]{fontenc}

\usepackage[utf8]{inputenc}
\usepackage{tabularx}
\usepackage{multirow}
\usepackage{booktabs}
\usepackage{graphicx}
\usepackage{subcaption}

\usepackage{microtype}

\usepackage{amsmath}
\usepackage{amssymb}
\usepackage{bbm}
\aclfinalcopy 
\title{Investigating the Reordering Capability in CTC-based Non-Autoregressive End-to-End Speech Translation}

\author{{\bf Shun-Po Chuang$^\ast$\quad Yung-Sung Chuang$^\ast$\quad Chih-Chiang Chang$^\ast$ \quad Hung-yi Lee}\\
National Taiwan University, Taipei, Taiwan \\
  \texttt{\{f04942141,b05901033,r09922057,hungyilee\}@ntu.edu.tw} \\}

\newcommand\blfootnote[1]{%
  \begingroup
  \renewcommand\thefootnote{}\footnote{#1}%
  \addtocounter{footnote}{-1}%
  \endgroup
}

\begin{document}
\maketitle

\blfootnote{$^\ast$Contributed equally.}
\begin{abstract}

We study the possibilities of building a non-autoregressive speech-to-text translation model using connectionist temporal classification (CTC), and use CTC-based automatic speech recognition as an auxiliary task to improve the performance.
CTC's success on translation is counter-intuitive due to its monotonicity assumption, so we analyze its reordering capability. 
Kendall's tau distance is introduced as the quantitative metric, and gradient-based visualization provides an intuitive way to take a closer look into the model. 
Our analysis shows that transformer encoders have the ability to change the word order and points out the future research direction that worth being explored more on non-autoregressive speech translation.\footnote{The source code is available. See Appendix~\ref{sec:appendix:code}.}

\end{abstract}

\section{Introduction}

Recently, there are more and more research works focusing on end-to-end speech translation (ST)~\citep{berard2016listen, weiss2017sequence, berard2018end, vila2018end, di2019adapting, ran2019guiding, chuang-etal-2020-worse}. Instead of cascading machine translation (MT) models to an automatic speech recognition (ASR) system, end-to-end models can skip the error bottleneck caused by ASR and be more computationally efficient. However, in the inference time, an autoregressive (AR) decoder is needed to decode the output sequence, causing the latency issue. 

In MT, non-autoregressive (NAR) models have been heavily explored recently~\citep{gu2018non, lee2018deterministic, ghazvininejad2019mask, stern2019insertion, gu2019levenshtein, saharia2020non} by leveraging the parallel nature of transformer~\citep{vaswani2017attention}.
In contrast, such kind of models is rarely explored in the field of speech translation, except for a concurrent work~\citep{inaguma2020orthros}. 
In this work, we use connectionist temporal classification (CTC)~\citep{graves2006connectionist} to train NAR models for ST, without an explicit decoder module. 
Our entire model is merely a transformer encoder. Multitask learning~\citep{anastasopoulos2018tied,9383496} on ASR, which is often used in speech translation, can also be applied in our transformer encoder architecture to further push the performance. 
We achieve initial results on NAR speech translation by using a single speech encoder. 

CTC's success on the translation task is counter-intuitive because of its monotonicity assumption. 
Previous works directly adopt the CTC loss on NAR translation without further verification on the reordering capability of CTC~\cite{libovicky2018end, saharia2020non, inaguma2020orthros}.
To further understand the reason that the CTC-based model can achieve ST task, we analyze the ordering capabilities of ST models by leveraging Kendall's tau distance~\cite{birch2011reordering, kendall1938new}, and a gradient-based visualization is introduced to provide additional evidence.
To the best of our knowledge, this is the first time to examine the ordering capabilities on the ST task.

We found that after applying multitask training, our model can have more tendency to re-arrange the positions of the target words to better positions that are not aligned with audio inputs. 
We highlight that our contribution is to 1) take the first step on translating pure speech signal to target language text in a NAR end-to-end manner and 2) take a closer look at the reason that NAR model with CTC loss can achieve non-monotonic mapping.

\section{Approaches}
\label{sec:approaches}
\subsection{CTC-based NAR-ST Model}
\label{subsec:nat_st_model}

We adopt transformer architecture for non-autoregressive speech-to-text translation (NAR-ST). 
The NAR-ST model consists of convolutional layers and self-attention layers.
The audio sequence $X$ is downsampled by convolutional layers, and self-attention layers will generate the final translation token sequence $Y$ based on the downsampled acoustic features. 
We use CTC loss as the objective function to optimize the NAR-ST model. 
The CTC decoding algorithm allows the model to generate translation in a single step.

CTC predicts an alignment between the input audio sequence $X$ and the output target sequence $Y$ by considering the probability distribution marginalized over all possible alignments.

The CTC loss function is defined as:
\begin{equation}
\mathcal{L}_{\mathrm{CTC}} = - \sum_{(x, y)\in D} \log p_\theta(y|x),
\end{equation}
where $x$ is audio frame sequence, $y$ is target sequence, and $D$ is the training set.
CTC uses dynamic programming to marginalize out the latent alignments to compute the log-likelihood:
\begin{equation}
\log p_\theta(y|x) = \log \sum_{a \in \beta(y)} \prod_{t} p(a_t|x;\theta)
\end{equation}
where $a = \{a_t\}_{t=0}^{|x|}$ is an alignment between $x$ and $y$ and is allowed to include a special ``blank'' token that should be removed when converting $a$ to the target sequence $y$. $\beta^{-1}(a)$ is a collapsing function such that $\beta^{-1}(a) = y$ if $a \in \beta(y)$.

CTC has a strong conditional independence and monotonicity assumption. It means that the tokens in $Y$ can be generated independently, and there exists a monotonic alignment between $X$ and $Y$. 
The monotonicity property is suitable for tasks such as ASR.
However, in translation tasks, there is no guarantee that the output sequence should follow the assumption, as word orders differ in different languages. 
In this work, we want to examine whether the powerful self-attention based transformer model can overcome this problem to some degree or not. 

\subsection{CTC-based Multitask NAR-ST Model}
\label{subsec:multitask_model}

Multitask learning improves data efficiency and performance across various tasks~\citep{zhang2017survey}. 
In AR end-to-end ST, multitask learning technique is often applied using ASR as an auxiliary task~\citep{anastasopoulos2018tied, sperber2020speech}. 
It requires an ASR decoder in addition to the ST decoder to learn to predict transcriptions while sharing the encoder. 

\begin{figure}[t!]
    \centering
    \includegraphics[width=\linewidth]{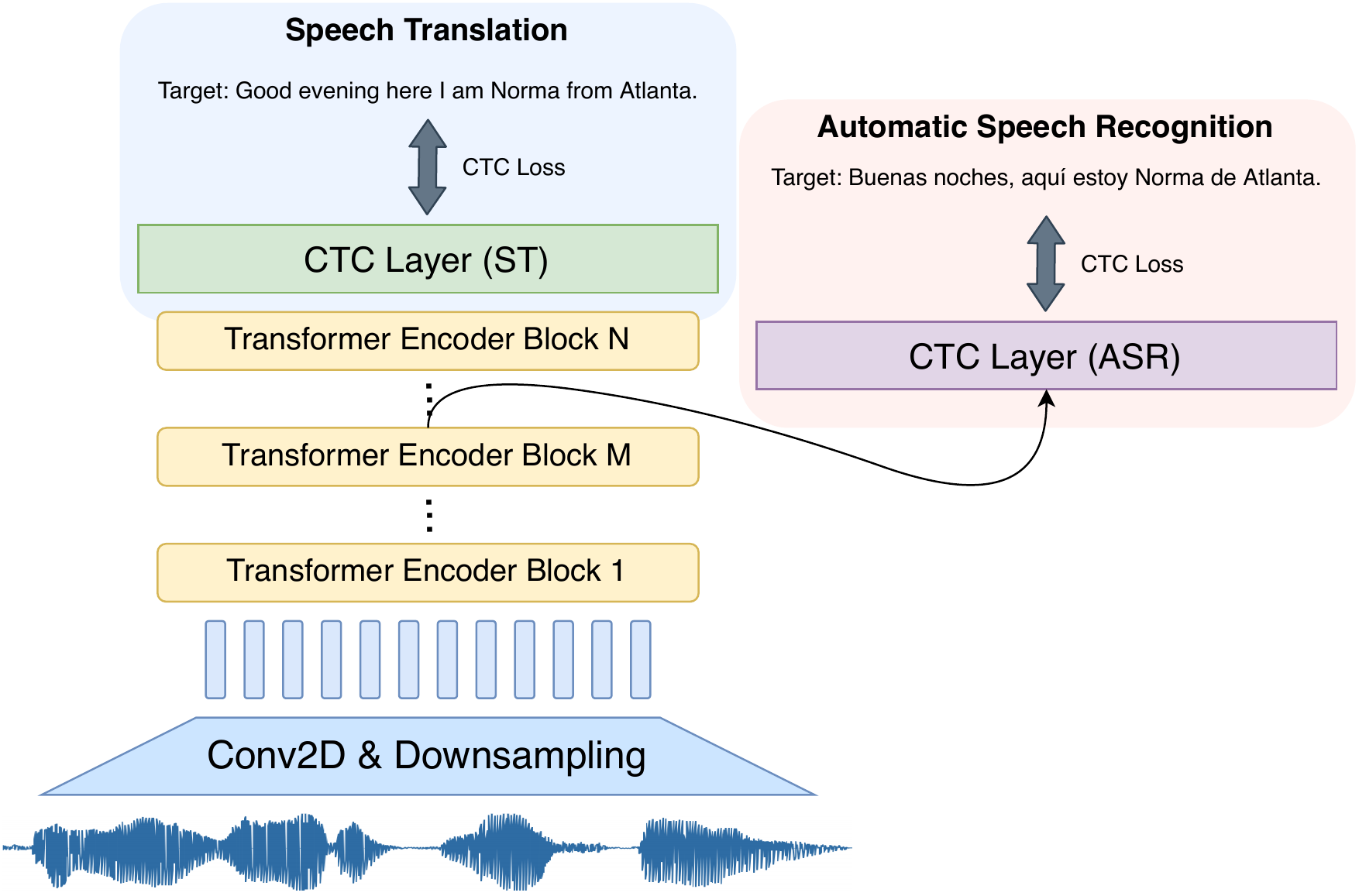}
    \caption{Multitask CTC model for ST and ASR.}
    \label{fig:ctc}
\end{figure}

To perform multitask learning on NAR-ST model, we propose to apply CTC-based ASR on a single $M$-th layer in the model, as illustrated in Figure~\ref{fig:ctc}.
It helps the NAR-ST model capture more information with a single CTC layer in an end-to-end manner.
And the ASR output will not be involved in the translation decoding process. 

\begin{table*}[ht!]
\small
\centering
\setlength\tabcolsep{3pt}
\begin{tabular}{cc|cccccc|c}
\toprule
\multicolumn{3}{c}{\bf Method} & \bf fisher\_dev	& \bf fisher\_dev2 & \bf fisher\_test & \bf CH\_devtest & \bf CH\_evltest & \bf Speed-Up\\
\midrule
(A) & \multicolumn{2}{l}{\textit{Autoregressive Models}} \\
\midrule
(a) & \multicolumn{2}{l}{Transformer (\textit{b}=10)} & 38.90 & 39.78 & 38.92 & 12.93 & 13.05 & $\times$ 1.0 \\
(b) & \multicolumn{2}{l}{Transformer + MTL (\textit{b}=10)} & 46.38 & 47.54 & 46.28 & \bf 17.66 & 17.17 & $\times$ 1.0 \\
(c) & \multicolumn{2}{l}{Transformer + MTL + ASR init. (\textit{b}=10)} & \bf 48.27 & \bf 49.17 & \bf 48.40 & 17.26 & 17.45 & $\times$ 1.0 \\
(d) & \multicolumn{2}{l}{Transformer + MTL + ASR init. (\textit{b}=5)} & 48.18 & 48.91 & 48.21 & 17.34 & \bf 17.55 & $\times$ 2.0 \\
(e) & \multicolumn{2}{l}{Transformer + MTL + ASR init. (\textit{b}=1)} & 46.05 & 47.04 & 46.14 & 16.48 & 16.33 & $\times$ 8.5 \\
\midrule
(B) & \multicolumn{2}{l}{\textit{Non-Autoregressive Models (Ours)}} \\
\midrule
(f) & \multicolumn{2}{l}{CTC} & 42.61 & 43.91 & 43.50 & 13.02 & 13.52 & \multirow{5}{*}{$\times$ 28.9} \\
(g) & \multicolumn{2}{l}{CTC + MTL at 4-th layer} & 42.26 & 43.70 & 43.58 & 13.10 & 13.17 & \multirow{5}{*}{}\\
(h) & \multicolumn{2}{l}{CTC + MTL at 6-th layer} & 42.06 & 44.05 & 43.56 & 13.19 & 13.38 & \multirow{5}{*}{}\\
(i) & \multicolumn{2}{l}{CTC + MTL at 8-th layer} & \bf 44.45 & \bf 45.23 & \bf 44.92 & \bf 14.20 & \bf 14.19 & \multirow{5}{*}{}\\
(j) & \multicolumn{2}{l}{CTC + MTL at 10-th layer} & 42.86 & 44.18 & 43.59 & 13.65 & 13.28 & \multirow{5}{*}{}\\

\bottomrule
\end{tabular}
\vspace{-5pt}
\caption{BLEU on Fisher Spanish dataset and CALLHOME (CH) dataset, including autoregressive and non-autoregressive models. The abbreviation \textit{b} stands for the beam size for beam search decoding. Multitask learning (MTL) represents using ASR as the auxiliary task trained with ST. In \textit{Autoregressive Models}, the auxiliary loss always applied on the final encoder output in MTL, and we applied it on different layers in NAR models.}
\label{tab:fisher_callhome}
\vspace{-5pt}
\end{table*}
\subsection{Reordering Evaluation -- Kendall's Tau Distance}
\label{sec:kendalltau_intro}
We measure reordering degree by Kendall's tau distance~\cite{kendall1938new}. LRscore~\cite{birch2011reordering} also introduced the distance with consideration of lexical correctness. 
Different from LRscore, we purely analyze the reordering capability rather than lexical correctness in this work.

Given a sentence triplet $(T, H, Y)$, where $T=\langle t_1,...,t_{|T|} \rangle$ is the audio transcription.
$H=\langle h_1,...,h_{|H|} \rangle$ and $Y=\langle y_1,...,y_{|Y|} \rangle$ are hypothesis and reference translation, respectively.
An external aligner provides two alignments:
$\boldsymbol{\pi}=\langle \pi(1),...,\pi(|T|) \rangle$ and $\boldsymbol{\sigma}=\langle \sigma(1),...,\sigma(|T|) \rangle$.
$\boldsymbol{\pi}$ maps each source token $t_k$ to a reference token $y_{\pi(k)}$, and $\boldsymbol{\sigma}$ maps $t_k$ to a hypothesis token $h_{\sigma(k)}$.
We follow the simplifications proposed in LRscore to reduce the alignments to a bijective relationship. 
The proportion of disagreements between $\boldsymbol{\pi}$ and $\boldsymbol{\sigma}$ is:
\begin{align*}
    \mathrm{R}(\boldsymbol{\pi}, \boldsymbol{\sigma}) &= \frac{\sum_{i=1}^{|T|}\sum_{j=1}^{|T|}z_{ij}}{|T|(|T|-1)/2} \\
    \text{where }z_{ij} &= \begin{cases} 
        1&\text{if $\pi(i)<\pi(j)$ and $\sigma(i)>\sigma(j)$} \\ 
        0&\text{otherwise} 
    \end{cases}
\end{align*}

Then, we define the term \textbf{reordering correctness} $\mathrm{R}_{acc}$ by introducing the brevity penalty ($BP$):
\begin{equation*}
\begin{split}
    \mathrm{R}_{acc} &= (1 -
    \sqrt{\mathrm{R}(\boldsymbol{\pi},
    \boldsymbol{\sigma}})) * BP\\
    \text{where }BP&=e^{1-|Y|/|H|}\text{ if $|H|\leq |Y|$ else }1
\end{split}
\end{equation*}
The higher the value, the more similar between two given alignments. 
Ideally, a well-trained model could handle the reordering problem by making $\boldsymbol{\sigma}$ close to $\boldsymbol{\pi}$ and result in $R_{acc}=1$.

\vspace{-3pt}
\section{Experiments}
\vspace{-4pt}
\subsection{Experimental Setup}
\label{subsec:exp_setup}

We use the ESPnet toolkit~\citep{watanabe2018espnet} for experiments.
We perform Spanish speech to English text translation with Fisher Spanish corpus. 
The test sets of CALLHOME corpus are also included for evaluation. The dataset details and download links are listed in Appendix~\ref{sec:appendix:data}. 

The NAR-ST model consists of two convolutional layers followed by twelve transformer encoder layers. 
Knowledge distillation~\citep{kim2016sequence} is also applied. More training details and parameter settings can be found in Appendix~\ref{sec:appendix:params}.

\subsection{Translation Quality and Speed}
\label{exp:bleu_eval}

We use BLEU~\citep{papineni2002bleu} to evaluate the translation quality, as shown in Table~\ref{tab:fisher_callhome}. Beam-search decoding with beam-size \textit{b} is considered for the AR models in this experiments. Greedy decoding is always used for the NAR models.

In the results of AR models (part (A)), multitask learning (MTL) can get better performance compared to the model without jointly training with an auxiliary task (row (b) v.s (a)).
Further improvement can be brought by using a pre-trained ASR encoder as the initialization weight (row (c) v.s (b)). It shows that using ASR data for MTL and initialization are the essential steps to achieve exceptional performance.
The performance drops when beam-size decreases, which shows a trade-off between the decoding speed and the performance (row (c) v.s (d)(e)). 

To better optimize the decoding speed, NAR-ST provide a great solution to reach a shorter decoding time (part (B)).
NAR-ST models is $\times$28.9 faster than the AR model with beam-size 10 (part (B) v.s rows (a)-(c)) and $\times$3.4 faster than the AR model with greedy decoding (part (B) v.s row (e)).
We initialize the NAR-ST models with the weight pre-trained on ASR task and applied the proposed MTL approach on different intermediate layers (rows (g)-(j)).
As the results showed in part (B),
applying MTL on the higher layers improves the performance (rows (i)(j) v.s (f)). 
It shows that speech signal needs more layers to model the complexity, and sheds light on selecting the intermediate layer to apply MTL is essential. We also evaluate the ASR results of the MTL models in Appendix~\ref{sec:appendix:wer}.

Some text-based refinement approaches can further improve the translation quality~\citep{libovicky2018end, lee2018deterministic, ghazvininejad2019mask, gu2019levenshtein, chan2020imputer}. We leave it as the future work and focus on analyzing the reordering capability of the CTC-based model.

\begin{table}[]
\centering
\small
\setlength\tabcolsep{6pt}
\begin{tabular}{@{}llccc}
\toprule
\multicolumn{2}{l}{}                         & \multicolumn{3}{c}{$\mathrm{R}_{acc}$}       \\ \midrule
\multicolumn{2}{c|}{\textbf{Method}}         & \textbf{dev} & \textbf{dev2} & \textbf{test} \\ \midrule
(a) & \multicolumn{1}{l|}{random permutaion} & 41.37        & 41.84         & 42.42         \\ \midrule
\multicolumn{5}{l}{\textit{Autoregress Models}}                                             \\ \midrule
(b) & \multicolumn{1}{l|}{Tr. (b=10)}        & 77.79        & 79.16         & 79.09         \\
(c) & \multicolumn{1}{l|}{Tr. + MTL (b=10)}   & 79.32        & 80.18         & 80.30         \\ \midrule
\multicolumn{5}{l}{\textit{Non-Autoregress Models (Ours)}}                                  \\ \midrule
(d) & \multicolumn{1}{l|}{CTC}               & 71.69        & 74.00         & 74.57         \\
(e) & \multicolumn{1}{l|}{CTC+MTL@8}         & 71.91        & 74.35         & 74.96         \\ \bottomrule
\end{tabular}
\caption{The reordering correctness evaluated on Fisher Spanish dataset. The values are selected as the best among 4 references. Tr. stands for Transformer. All models are initialized with pretrained ASR.}
\vspace{-5pt}
\label{tab:reorder}
\end{table}

\vspace{-3pt}
\subsection{Word Order Analysis}
\vspace{-3pt}
\label{exp:word_order}

In this section, we discuss the word ordering problem in the translation task.
We use $\mathrm{R}_{acc}$ defined in section~\ref{sec:kendalltau_intro} to systematically evaluate the reordering correctness across the corpora. 
Besides, we examine the gradient norm in models to visualize the reordering process. 

\paragraph{Quantitative Analysis}
We use SimAlign~\citep{sabet2020simalign} to align the transcriptions and translations with details in Appendix~\ref{sec:appendix:simalign}.
Table~\ref{tab:reorder} shows $\mathrm{R}_{acc}$ evaluated on ST models. We included the correctness of random permutation as a baseline.

The AR models obtain high $\mathrm{R}_{acc}$ scores (rows (b)(c)), it shows that the AR model can handle a complex word order. 
The NAR models also have the ability to rearrange words 
(rows (d)(e) v.s. (a))
but are weaker than AR models due to the independent assumption brought by
CTC. An interesting observation is that applying MTL tends to improve $\mathrm{R}_{acc}$
(rows (c)(e) v.s. (b)(d)). We conclude that the monotonic natural in ASR improves the stability in training ST~\cite{Sperber2019}.

\begin{figure}[t!]
    \centering
    \includegraphics[width=\linewidth]{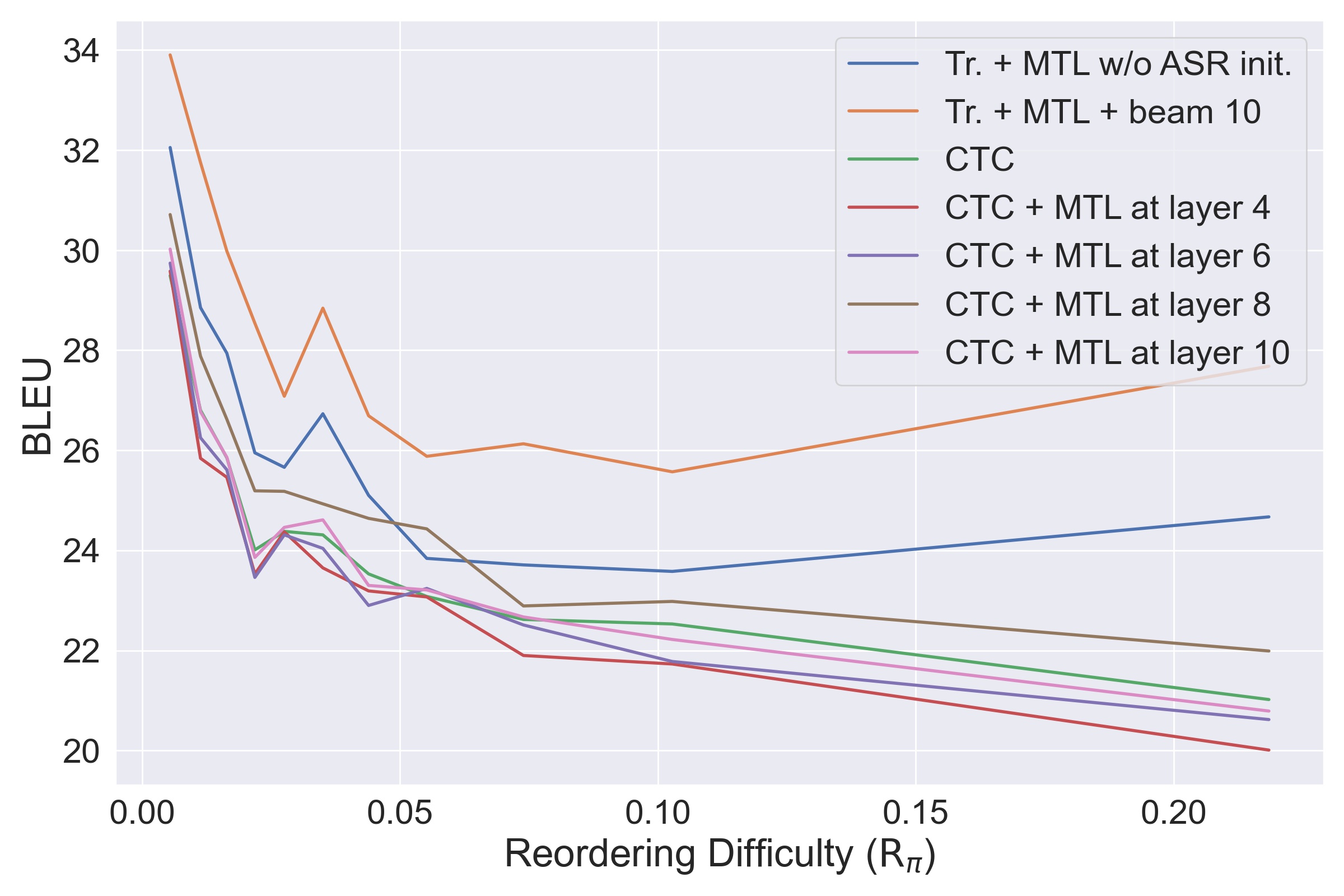}
    \caption{The BLEU score curve under different reordering difficulties ($\mathrm{R}_\pi$). Details are in Appendix~\ref{sec:appendix:bleu_ktau_curve_details}.}
    \label{fig:bleu_ktau_curve}
    \vspace{-5pt}
\end{figure}

To investigate the relation between model performance and the reordering difficulty, 
we measure the reordering difficulty by $\mathrm{R_{\boldsymbol{\pi}}}=\mathrm{R}(\boldsymbol{\pi}, \boldsymbol{m})$, where $\boldsymbol{m}=\langle 1,...,|T| \rangle$ is a dummy monotonic alignment.
We split all the testing data (dev/dev2/test) into smaller equal-size groups by different reference $\mathrm{R_{\boldsymbol{\pi}}}$. The BLEU scores for these groups were plotted in Figure~\ref{fig:bleu_ktau_curve}. Obviously, AR models are more robust to higher reordering difficulty. Nonetheless, we observed that when MTL is applied at layer 8, CTC model is more robust to reordering difficulty, in some cases ($\mathrm{R_{\boldsymbol{\pi}}}$<0.07) even come close to the AR model without ASR pretraining.

\paragraph{Gradient-based Visualization}
We consider the gradient norm as an approximated indicator of reordering in our model. 
For each output token $h_i$, we concatenate the relative influence on it across all layers, which yields a matrix $\mathbf{O}^i\in \mathbb{R}^{|X|\times L}$, where each row is a frame and each column is a layer. We refer to this as the reordering matrix for token $h_i$. We leave the computational details in Appendix~\ref{sec:appendix:gbv}.

\begin{figure}[t!]
    \centering
    \includegraphics[width=1.0\linewidth]{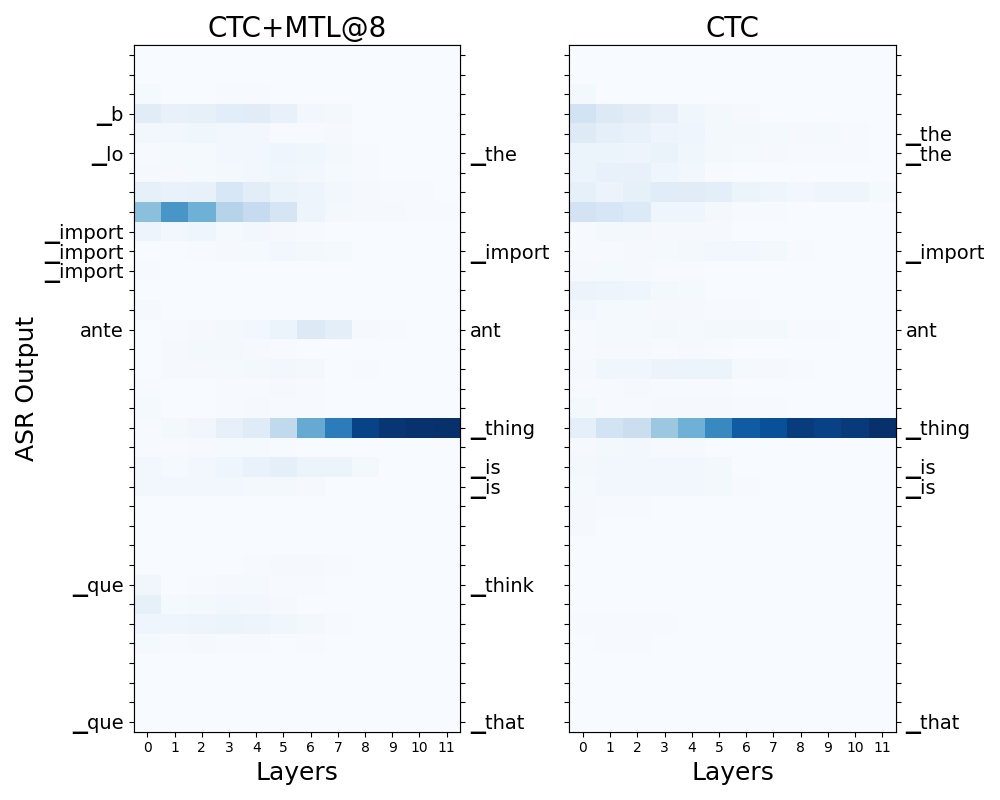}
    \vspace{-20pt}
    \caption{The re-ordering matrix of the token "thing". The ASR output from MTL model is included as a proxy for word positioning in source speech. }
    \vspace{-15pt}
    \label{fig:reorder_matrix}
\end{figure}

Figure~\ref{fig:reorder_matrix} shows a reordering matrix for token \texttt{\_thing}.
We can observe that the single-task CTC model (Figure~\ref{fig:reorder_matrix}, right) tends to keep focusing on the same position of the output token  \texttt{\_thing} at the lower layers. In contrast, the model with MTL (Figure~\ref{fig:reorder_matrix}, left) does not focus on the same position at the lower layers. It can focus more on the other position where the corresponding speech segment appears, and aggregate the information to the output position at the higher layers. Additional examples can be found in Appendix~\ref{sec:appendix:reorder}.

\vspace{-5pt}
\section{Concluding Remarks}
\vspace{-5pt}
\label{sec:conclusion}

We propose a CTC-based NAR-ST model with an auxiliary CTC-based ASR task
and are the first to study the reordering capability in CTC-based NAR-ST model. 
$\mathrm{R}_{acc}$ is adopted to analyze reordering in the ST task, and gradient-based visualizations reveal the internal manipulation of the models. Besides trying to improve BLEU scores, we encourage future research on NAR models to also evaluate whether the NAR models have inferior reordering capabilities in order to close the gap between AR and NAR models.

\section*{Broader Impact and Ethical Considerations}
We believe that our work can help researchers in the NLP community understand more about the non-autoregressive speech translation models, and we envision that the model proposed in this paper will equip the researchers with a new technique to perform better and faster speech translation. We do not see ourselves violating the code of ethics of ACL-IJCNLP 2021.

\bibliography{acl2021}

\begin{thebibliography}{32}
\expandafter\ifx\csname natexlab\endcsname\relax\def\natexlab#1{#1}\fi

\bibitem[{Anastasopoulos and Chiang(2018)}]{anastasopoulos2018tied}
Antonios Anastasopoulos and David Chiang. 2018.
\newblock Tied multitask learning for neural speech translation.
\newblock In \emph{Proceedings of the 2018 Conference of the North American
  Chapter of the Association for Computational Linguistics: Human Language
  Technologies, Volume 1 (Long Papers)}, pages 82--91.

\bibitem[{B{\'e}rard et~al.(2018)B{\'e}rard, Besacier, Kocabiyikoglu, and
  Pietquin}]{berard2018end}
Alexandre B{\'e}rard, Laurent Besacier, Ali~Can Kocabiyikoglu, and Olivier
  Pietquin. 2018.
\newblock End-to-end automatic speech translation of audiobooks.
\newblock In \emph{2018 IEEE International Conference on Acoustics, Speech and
  Signal Processing (ICASSP)}, pages 6224--6228. IEEE.

\bibitem[{B{\'e}rard et~al.(2016)B{\'e}rard, Pietquin, Servan, and
  Besacier}]{berard2016listen}
Alexandre B{\'e}rard, Olivier Pietquin, Christophe Servan, and Laurent
  Besacier. 2016.
\newblock Listen and translate: A proof of concept for end-to-end
  speech-to-text translation.
\newblock \emph{arXiv preprint arXiv:1612.01744}.

\bibitem[{Birch and Osborne(2011)}]{birch2011reordering}
Alexandra Birch and Miles Osborne. 2011.
\newblock Reordering metrics for mt.
\newblock In \emph{Proceedings of the 49th Annual Meeting of the Association
  for Computational Linguistics: Human Language Technologies}, pages
  1027--1035.

\bibitem[{Chan et~al.(2020)Chan, Saharia, Hinton, Norouzi, and
  Jaitly}]{chan2020imputer}
William Chan, Chitwan Saharia, Geoffrey Hinton, Mohammad Norouzi, and Navdeep
  Jaitly. 2020.
\newblock Imputer: Sequence modelling via imputation and dynamic programming.
\newblock \emph{arXiv preprint arXiv:2002.08926}.

\bibitem[{Chuang et~al.(2020)Chuang, Sung, Liu, and
  Lee}]{chuang-etal-2020-worse}
Shun-Po Chuang, Tzu-Wei Sung, Alexander~H. Liu, and Hung-yi Lee. 2020.
\newblock \href {https://doi.org/10.18653/v1/2020.acl-main.533} {Worse {WER},
  but better {BLEU}? leveraging word embedding as intermediate in multitask
  end-to-end speech translation}.
\newblock In \emph{Proceedings of the 58th Annual Meeting of the Association
  for Computational Linguistics}, pages 5998--6003, Online. Association for
  Computational Linguistics.

\bibitem[{Conneau et~al.(2019)Conneau, Khandelwal, Goyal, Chaudhary, Wenzek,
  Guzm{\'a}n, Grave, Ott, Zettlemoyer, and Stoyanov}]{conneau2019unsupervised}
Alexis Conneau, Kartikay Khandelwal, Naman Goyal, Vishrav Chaudhary, Guillaume
  Wenzek, Francisco Guzm{\'a}n, Edouard Grave, Myle Ott, Luke Zettlemoyer, and
  Veselin Stoyanov. 2019.
\newblock Unsupervised cross-lingual representation learning at scale.
\newblock \emph{arXiv preprint arXiv:1911.02116}.

\bibitem[{Di~Gangi et~al.(2019)Di~Gangi, Negri, and Turchi}]{di2019adapting}
Mattia~A Di~Gangi, Matteo Negri, and Marco Turchi. 2019.
\newblock Adapting transformer to end-to-end spoken language translation.
\newblock \emph{Proc. Interspeech 2019}, pages 1133--1137.

\bibitem[{Ghazvininejad et~al.(2019)Ghazvininejad, Levy, Liu, and
  Zettlemoyer}]{ghazvininejad2019mask}
Marjan Ghazvininejad, Omer Levy, Yinhan Liu, and Luke Zettlemoyer. 2019.
\newblock Mask-predict: Parallel decoding of conditional masked language
  models.
\newblock In \emph{Proceedings of the 2019 Conference on Empirical Methods in
  Natural Language Processing and the 9th International Joint Conference on
  Natural Language Processing (EMNLP-IJCNLP)}, pages 6114--6123.

\bibitem[{Graves et~al.(2006)Graves, Fern{\'a}ndez, Gomez, and
  Schmidhuber}]{graves2006connectionist}
Alex Graves, Santiago Fern{\'a}ndez, Faustino Gomez, and J{\"u}rgen
  Schmidhuber. 2006.
\newblock Connectionist temporal classification: labelling unsegmented sequence
  data with recurrent neural networks.
\newblock In \emph{Proceedings of the 23rd international conference on Machine
  learning}, pages 369--376.

\bibitem[{Gu et~al.(2018)Gu, Bradbury, Xiong, Li, and Socher}]{gu2018non}
Jiatao Gu, James Bradbury, Caiming Xiong, Victor~OK Li, and Richard Socher.
  2018.
\newblock Non-autoregressive neural machine translation.
\newblock In \emph{International Conference on Learning Representations}.

\bibitem[{Gu et~al.(2019)Gu, Wang, and Zhao}]{gu2019levenshtein}
Jiatao Gu, Changhan Wang, and Junbo Zhao. 2019.
\newblock Levenshtein transformer.
\newblock In \emph{Advances in Neural Information Processing Systems}, pages
  11181--11191.

\bibitem[{Inaguma et~al.(2020)Inaguma, Higuchi, Duh, Kawahara, and
  Watanabe}]{inaguma2020orthros}
Hirofumi Inaguma, Yosuke Higuchi, Kevin Duh, Tatsuya Kawahara, and Shinji
  Watanabe. 2020.
\newblock Orthros: Non-autoregressive end-to-end speech translation with
  dual-decoder.
\newblock \emph{arXiv preprint arXiv:2010.13047}.

\bibitem[{Kano et~al.(2021)Kano, Sakti, and Nakamura}]{9383496}
Takatomo Kano, Sakriani Sakti, and Satoshi Nakamura. 2021.
\newblock \href {https://doi.org/10.1109/SLT48900.2021.9383496}
  {Transformer-based direct speech-to-speech translation with transcoder}.
\newblock In \emph{2021 IEEE Spoken Language Technology Workshop (SLT)}, pages
  958--965.

\bibitem[{Kendall(1938)}]{kendall1938new}
Maurice~G Kendall. 1938.
\newblock A new measure of rank correlation.
\newblock \emph{Biometrika}, 30(1/2):81--93.

\bibitem[{Kim and Rush(2016)}]{kim2016sequence}
Yoon Kim and Alexander~M Rush. 2016.
\newblock Sequence-level knowledge distillation.
\newblock In \emph{Proceedings of the 2016 Conference on Empirical Methods in
  Natural Language Processing}, pages 1317--1327.

\bibitem[{Lee et~al.(2018)Lee, Mansimov, and Cho}]{lee2018deterministic}
Jason Lee, Elman Mansimov, and Kyunghyun Cho. 2018.
\newblock Deterministic non-autoregressive neural sequence modeling by
  iterative refinement.
\newblock In \emph{Proceedings of the 2018 Conference on Empirical Methods in
  Natural Language Processing}, pages 1173--1182.

\bibitem[{Libovick{\`y} and Helcl(2018)}]{libovicky2018end}
Jind{\v{r}}ich Libovick{\`y} and Jind{\v{r}}ich Helcl. 2018.
\newblock End-to-end non-autoregressive neural machine translation with
  connectionist temporal classification.
\newblock In \emph{Proceedings of the 2018 Conference on Empirical Methods in
  Natural Language Processing}, pages 3016--3021.

\bibitem[{Papineni et~al.(2002)Papineni, Roukos, Ward, and
  Zhu}]{papineni2002bleu}
Kishore Papineni, Salim Roukos, Todd Ward, and Wei-Jing Zhu. 2002.
\newblock Bleu: a method for automatic evaluation of machine translation.
\newblock In \emph{Proceedings of the 40th annual meeting of the Association
  for Computational Linguistics}, pages 311--318.

\bibitem[{Park et~al.(2019)Park, Chan, Zhang, Chiu, Zoph, Cubuk, and
  Le}]{park2019specaugment}
Daniel~S Park, William Chan, Yu~Zhang, Chung-Cheng Chiu, Barret Zoph, Ekin~D
  Cubuk, and Quoc~V Le. 2019.
\newblock Specaugment: A simple data augmentation method for automatic speech
  recognition.
\newblock \emph{Proc. Interspeech 2019}, pages 2613--2617.

\bibitem[{Ran et~al.(2019)Ran, Lin, Li, and Zhou}]{ran2019guiding}
Qiu Ran, Yankai Lin, Peng Li, and Jie Zhou. 2019.
\newblock Guiding non-autoregressive neural machine translation decoding with
  reordering information.
\newblock \emph{arXiv preprint arXiv:1911.02215}.

\bibitem[{Sabet et~al.(2020)Sabet, Dufter, and Sch{\"u}tze}]{sabet2020simalign}
Masoud~Jalili Sabet, Philipp Dufter, and Hinrich Sch{\"u}tze. 2020.
\newblock Simalign: High quality word alignments without parallel training data
  using static and contextualized embeddings.
\newblock \emph{arXiv preprint arXiv:2004.08728}.

\bibitem[{Saharia et~al.(2020)Saharia, Chan, Saxena, and
  Norouzi}]{saharia2020non}
Chitwan Saharia, William Chan, Saurabh Saxena, and Mohammad Norouzi. 2020.
\newblock Non-autoregressive machine translation with latent alignments.
\newblock \emph{arXiv preprint arXiv:2004.07437}.

\bibitem[{Sennrich et~al.(2016)Sennrich, Haddow, and
  Birch}]{sennrich2016neural}
Rico Sennrich, Barry Haddow, and Alexandra Birch. 2016.
\newblock Neural machine translation of rare words with subword units.
\newblock In \emph{Proceedings of the 54th Annual Meeting of the Association
  for Computational Linguistics (Volume 1: Long Papers)}, pages 1715--1725.

\bibitem[{Sperber et~al.(2019)Sperber, Neubig, Niehues, and
  Waibel}]{Sperber2019}
Matthias Sperber, Graham Neubig, Jan Niehues, and Alex Waibel. 2019.
\newblock \href {https://arxiv.org/abs/1904.07209} {{Attention-Passing Models
  for Robust and Data-Efficient End-to-End Speech Translation}}.
\newblock \emph{Transactions of the Association for Computational Linguistics
  (TACL)}.

\bibitem[{Sperber and Paulik(2020)}]{sperber2020speech}
Matthias Sperber and Matthias Paulik. 2020.
\newblock Speech translation and the end-to-end promise: Taking stock of where
  we are.
\newblock \emph{arXiv preprint arXiv:2004.06358}.

\bibitem[{Stern et~al.(2019)Stern, Chan, Kiros, and
  Uszkoreit}]{stern2019insertion}
Mitchell Stern, William Chan, Jamie Kiros, and Jakob Uszkoreit. 2019.
\newblock Insertion transformer: Flexible sequence generation via insertion
  operations.
\newblock In \emph{International Conference on Machine Learning}, pages
  5976--5985.

\bibitem[{Vaswani et~al.(2017)Vaswani, Shazeer, Parmar, Uszkoreit, Jones,
  Gomez, Kaiser, and Polosukhin}]{vaswani2017attention}
Ashish Vaswani, Noam Shazeer, Niki Parmar, Jakob Uszkoreit, Llion Jones,
  Aidan~N Gomez, {\L}ukasz Kaiser, and Illia Polosukhin. 2017.
\newblock Attention is all you need.
\newblock In \emph{Advances in neural information processing systems}, pages
  5998--6008.

\bibitem[{Vila et~al.(2018)Vila, Escolano, Fonollosa, and
  Costa-Juss{\`a}}]{vila2018end}
Laura-Cross Vila, Carlos Escolano, Jos{\'e}~AR Fonollosa, and Marta-R
  Costa-Juss{\`a}. 2018.
\newblock End-to-end speech translation with the transformer.
\newblock \emph{Proc. IberSPEECH 2018}, pages 60--63.

\bibitem[{Watanabe et~al.(2018)Watanabe, Hori, Karita, Hayashi, Nishitoba,
  Unno, Soplin, Heymann, Wiesner, Chen et~al.}]{watanabe2018espnet}
Shinji Watanabe, Takaaki Hori, Shigeki Karita, Tomoki Hayashi, Jiro Nishitoba,
  Yuya Unno, Nelson-Enrique~Yalta Soplin, Jahn Heymann, Matthew Wiesner, Nanxin
  Chen, et~al. 2018.
\newblock Espnet: End-to-end speech processing toolkit.
\newblock \emph{Proc. Interspeech 2018}, pages 2207--2211.

\bibitem[{Weiss et~al.(2017)Weiss, Chorowski, Jaitly, Wu, and
  Chen}]{weiss2017sequence}
Ron~J Weiss, Jan Chorowski, Navdeep Jaitly, Yonghui Wu, and Zhifeng Chen. 2017.
\newblock Sequence-to-sequence models can directly translate foreign speech.
\newblock \emph{arXiv preprint arXiv:1703.08581}.

\bibitem[{Zhang and Yang(2017)}]{zhang2017survey}
Yu~Zhang and Qiang Yang. 2017.
\newblock A survey on multi-task learning.
\newblock \emph{arXiv preprint arXiv:1707.08114}.

\end{thebibliography}
\bibliographystyle{acl_natbib}
\clearpage
\newpage
\appendix

\section{Source Code}
\label{sec:appendix:code}

Please download our code at \url{https://github.com/voidism/NAR-ST} and follow the instructions written in \texttt{README.md} to reproduce the results.

\section{Dataset Details}
\label{sec:appendix:data}

We use the Fisher and CALLHOME Spanish dataset (a Spanish-to-English speech-to-text translation dataset), which can be downloaded in the following links: 1) Fisher Spanish Speech \url{https://catalog.ldc.upenn.edu/LDC2010S01} 2) CALLHOME Spanish Speech \url{https://catalog.ldc.upenn.edu/LDC96S35} 3) 
Fisher and CALLHOME Spanish--English Speech Translation \url{https://catalog.ldc.upenn.edu/LDC2014T23}.

\subsection{Statistics}

The data statistics are listed in Table~\ref{tab:data}.

\begin{table}[h!]
    \centering
    \small
    \begin{tabular}{lcc}
        \toprule
        \bf Data Split & \bf \# of Utterance & \bf Duration (hours) \\
        \midrule
        \multicolumn{2}{l}{\it Training sets} & \\
        \midrule
        fisher\_train & 138792 & 171.61 \\
        \midrule
        \multicolumn{2}{l}{\it Validation sets} & \\
        \midrule
        fisher\_dev & 3973 & 4.59 \\
        fisher\_dev2 & 3957 & 4.70 \\
        callhome\_devtest & 3956 & 3.82 \\
        \midrule
        \multicolumn{2}{l}{\it Testing sets} & \\
        \midrule
        fisher\_test & 3638 & 4.48 \\
        callhome\_evltest & 1825 & 1.83 \\
        \bottomrule
    \end{tabular}
    \caption{The data statistics of the Fisher and CALLHOME Spanish dataset.}
    \label{tab:data}
\end{table}

\subsection{Preprocessing}
We use ESPnet to preprocess our data. For text, we use Byte Pair Encoding (BPE)~\citep{sennrich2016neural} with vocabulary size 8,000. We convert all text to lowercase with punctuation removal. For audio, we convert all audio files into wav file with a sample frequency of 16,000. We extract 80-dim fbank without delta. We use SpecAugment~\citep{park2019specaugment} to augment our data. More details can be found in our source code in Appendix~\ref{sec:appendix:code}.

\section{Training Details}
\label{sec:appendix:params}

\subsection{Computing Infrastructure and Runtime}
We use a single NVIDIA TITAN RTX (24G) for each experiment. The average runtime of experiments in Table~\ref{tab:fisher_callhome} is 2-3 days for both autoregressive and non-autoregressive models. 

\subsection{Hyperparameters}
Our training hyperparameters are listed in Table~\ref{tab:params}. We do not conduct hyperparameter search, but follow the autoregressive ST best setting in ESPnet toolkit~\citep{watanabe2018espnet}, and use the same hyperparameter for our non-autoregressive models. Due to the limited budget, we run each experiment for once. For inference stage of CTC-based models, we simply use greedy decode to produce the output sequences.

\begin{table}[h!]
    \centering
    \small
    \begin{tabular}{cc}
        \toprule
        \bf Hyperparameter & \bf Value \\
        \midrule
        encoder layers & 12 \\
        hidden units & 2048 \\
        attention dimension & 256 \\
        attention heads & 4 \\
        label smoothing weight & 0.1 \\
        batch size & 64 \\
        optimizer & noam \\
        learning rate & 2.5 \\
        warmup steps & 25000 \\
        attention dropout rate & 0.0 \\
        gradient accumulate step & 2 \\
        gradient clipping & 5.0 \\
        epoch & 30 \\
        dropout rate & 0.1 \\
        \bottomrule
    \end{tabular}
    \caption{The main hyperparameters in the experiment.}
    \label{tab:params}
\end{table}

\subsection{Knowledge Distillation}
To perform sequence-level knowledge distillation (Seq-KD)~\citep{kim2016sequence} to improve the performance of NAR models, we firstly trained an autoregressive transformer-based MT model on the transcriptions and translations in same training set with ESPnet.
Then we used the trained model to produce the hypotheses with beam search size of 1 for the whole training set. We swapped the ground truth sequences with the hypotheses for all NAR ST model training. We also show the ablation results on knowledge distillation in Table~\ref{tab:kd}.

We also try the possibility of using the autoregressive ST model to produce the hypotheses for Seq-KD, but the results are not as good as using a MT model. The results are shown in the second row in Table~\ref{tab:kd}. The download links to MT/ST decode results for conducting Seq-KD are also provided in the \texttt{README.md} file in our source code (See Appendix~\ref{sec:appendix:code}).

\begin{table}[ht!]
\small
\centering
\setlength\tabcolsep{3pt}
\begin{tabular}{c|ccc|cc}
\toprule
\multirow{2}{*}{\bf Method} & \multicolumn{3}{c|}{\bf Fisher} & \multicolumn{2}{c}{\bf CALLHOME} \\
\multirow{2}{*}{} & \bf dev	& \bf dev2 & \bf test & \bf devtest & \bf evltest \\
\midrule
\multicolumn{1}{l|}{NAR w/ MT Seq-KD} & \bf 44.45 & \bf 45.23 & \bf 44.92 & \bf 14.20 & \bf 14.19 \\
\multicolumn{1}{l|}{NAR w/ ST Seq-KD} & 41.08 & 41.28 & 40.85 & 13.63 & 13.27 \\
\multicolumn{1}{l|}{NAR w/o Seq-KD} & 41.59 & 43.13 & 42.25 & 13.05 & 12.81 \\
\bottomrule
\end{tabular}
\vspace{-5pt}
\caption{Ablation test on sequence-level knowledge distillation (Seq-KD) on Fisher Spanish dataset and CALLHOME (CH) dataset. The NAR model we used is CTC+MTL at 8-th layer.}
\label{tab:kd}
\vspace{-5pt}
\end{table}

\subsection{Model Selection}
When evaluating the models, we average the model checkpoints of the final 10 epochs to obtain our final model for NAR experiments. For AR experiments, we follow the original setting in ESPnet to average the 5 best-validated model checkpoints to obtain the final model.

\subsection{Model Size}

The number of parameters of our CTC model is 18.2M. The number of parameters of the autoregressive model is 27.9M. 

\section{Automatic Speech Recognition Evaluation}
\label{sec:appendix:wer}

\begin{table}[ht!]
\small
\centering
\setlength\tabcolsep{3pt}
\begin{tabular}{c|ccc}
\toprule
\multirow{2}{*}{\bf Method} & \multicolumn{3}{c}{\bf Fisher} \\
\multirow{2}{*}{} & \bf dev & \bf dev2 & \bf test  \\
\midrule
\multicolumn{1}{l|}{CTC+MTL@4} & 47.84 & 46.79 & 45.97 \\
\multicolumn{1}{l|}{CTC+MTL@6} & 40.21 & 39.02 & 38.08 \\
\multicolumn{1}{l|}{CTC+MTL@8} & 32.87 & 31.99 & 30.88 \\
\multicolumn{1}{l|}{CTC+MTL@10} & \bf 29.31 & \bf 28.36 & \bf 27.10 \\
\bottomrule
\end{tabular}
\vspace{-5pt}
\caption{The Word Error Rate (WER) on Fisher Spanish dataset of CTC results of the intermediate layers in multitask CTC models.}
\label{tab:wer}
\vspace{-5pt}
\end{table}

We compute the Word Error Rate (WER) for ASR output obtained from the intermediate ASR branch of our proposed models. The results are shown in Table~\ref{tab:wer}. We can observe that when applying multitask learning in the higher layers, the WER becomes lower. It indicates that ASR need more layers to perform better. However, the best ST scores are achieved by CTC+MTL@8 instead of CTC+MTL@10. It may be caused by the fact that there are only two transformer encoder layers for CTC+MTL@10 to perform ST. It may be too difficult for the model to convert the information from source language to target language in two encoder layers, even though the lower WER indicates useful information is provided to perform ST.

\section{SimAlign setup}
\label{sec:appendix:simalign}
We use the released code\footnote{\url{https://github.com/cisnlp/simalign}} by the authors of SimAlign as the external aligner to obtain word alignments used for calculating reordering metrics. SimAlign uses contextualized embeddings from pretrained language models, and there are several proposed algorithms to do word alignments. We use XLM-R~\citep{conneau2019unsupervised} as the underlying contextualized word embeddings with the itermax matching algorithm.

\section{Reordering Difficulty}
\label{sec:appendix:reorder_covost}
We provide reordering difficulty measured on all en-xx language pairs in CoVoST2 dataset in Table~\ref{tab:covost-difficulty}.
\begin{table}[h!]
\centering
\small
\begin{tabular}{@{}lrlr@{}}
\toprule
en-xx                   & \multicolumn{1}{c}{$\mathrm{R}_\pi$} & en-xx                   & \multicolumn{1}{c}{$\mathrm{R}_\pi$} \\ \midrule
\multicolumn{1}{l|}{de} & \multicolumn{1}{r|}{5.92} & \multicolumn{1}{l|}{tr} & 15.80                     \\
\multicolumn{1}{l|}{zh-CN} & \multicolumn{1}{r|}{10.63} & \multicolumn{1}{l|}{fa}    & 13.93 \\
\multicolumn{1}{l|}{ja}    & \multicolumn{1}{r|}{20.86} & \multicolumn{1}{l|}{sv-SE} & 2.94  \\
\multicolumn{1}{l|}{ar} & \multicolumn{1}{r|}{6.04} & \multicolumn{1}{l|}{mn} & 16.28                     \\
\multicolumn{1}{l|}{et} & \multicolumn{1}{r|}{5.98} & \multicolumn{1}{l|}{cy} & 5.78                      \\
\multicolumn{1}{l|}{ca} & \multicolumn{1}{r|}{4.39} & \multicolumn{1}{l|}{id} & 3.41                      \\
\multicolumn{1}{l|}{sl} & \multicolumn{1}{r|}{5.13} & \multicolumn{1}{l|}{ta} & 15.57                     \\
\multicolumn{1}{l|}{lv} & \multicolumn{1}{r|}{5.34} & \multicolumn{1}{l|}{}   &                           \\ \bottomrule
\end{tabular}
\caption{The training set reordering difficulty evaluated on each pairs of languages in the CoVoST2 dataset.}
\label{tab:covost-difficulty}
\end{table}

\section{Details on Figure~\ref{fig:bleu_ktau_curve}}
\label{sec:appendix:bleu_ktau_curve_details}
In Figure~\ref{fig:bleu_ktau_curve}, the primary goal is to view the relation between reordering difficulty and the model's performance. We describe the method used in~\citep{birch2011reordering} to represent the reordering difficulty as follows: For each example in the fisher dev and test set, calculate Kendall's tau distance between 1) its reference alignment (alignment between source transcription and reference translation) and 2) a dummy monotonic alignment, which is just the sequence 1...m. Intuitively this shows how much the reference alignment disagrees with monotonic alignment, and hence the reordering difficulty. Next, we divide all examples into 10 bins, where each bin contains examples with similar reordering difficulty, and all bins have an equal number of examples. Finally, we calculate the BLEU score of the hypotheses in each bin. The result is plotted in Figure~\ref{fig:bleu_ktau_curve}.

\section{Gradient-based Visualization}
\label{sec:appendix:gbv}
\begin{figure}[h!]
    \centering
    \includegraphics[width=\linewidth]{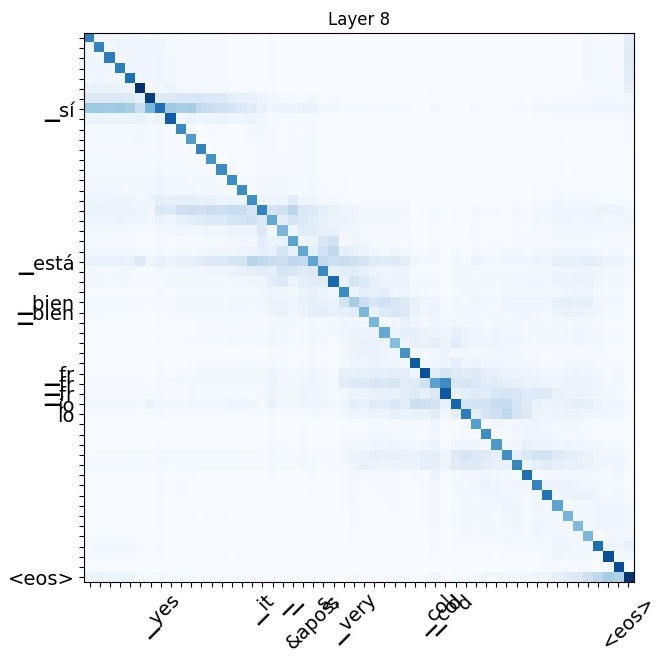}
    \caption{The saliency matrix of layer 8 input.}
    \label{fig:saliency_frio}
\end{figure}

We first obtain a saliency matrix $\mathbf{J}^M\in \mathbb{R}^{|X|\times|X|}$ for the $M$-th transformer layer by computing the gradient norm of output logits w.r.t. the latent representations of each timestep in that layer. An example is shown in Figure~\ref{fig:saliency_frio}.
Then, we normalize $\mathbf{J}^M$ across the dimension corresponding to the source audio sequence. 
Intuitively, the $i$-th column of $\mathbf{J}^M$ can be interpreted as the relative influence of the representations at each position on the $i$-th output token. 

Consequently, we proceed to re-arrange $\mathbf{J}^M$ in the following way: for each output token $h_i$, we concatenate the relative influence on it across all layers, which yields the reordering matrix for token $h_i$, denoted as $\mathbf{O}^i\in \mathbb{R}^{|X|\times L}$.

\section{Reordering Matrix}
\label{sec:appendix:reorder}
\begin{figure}[ht!]
    \centering
    \includegraphics[width=1.0\linewidth]{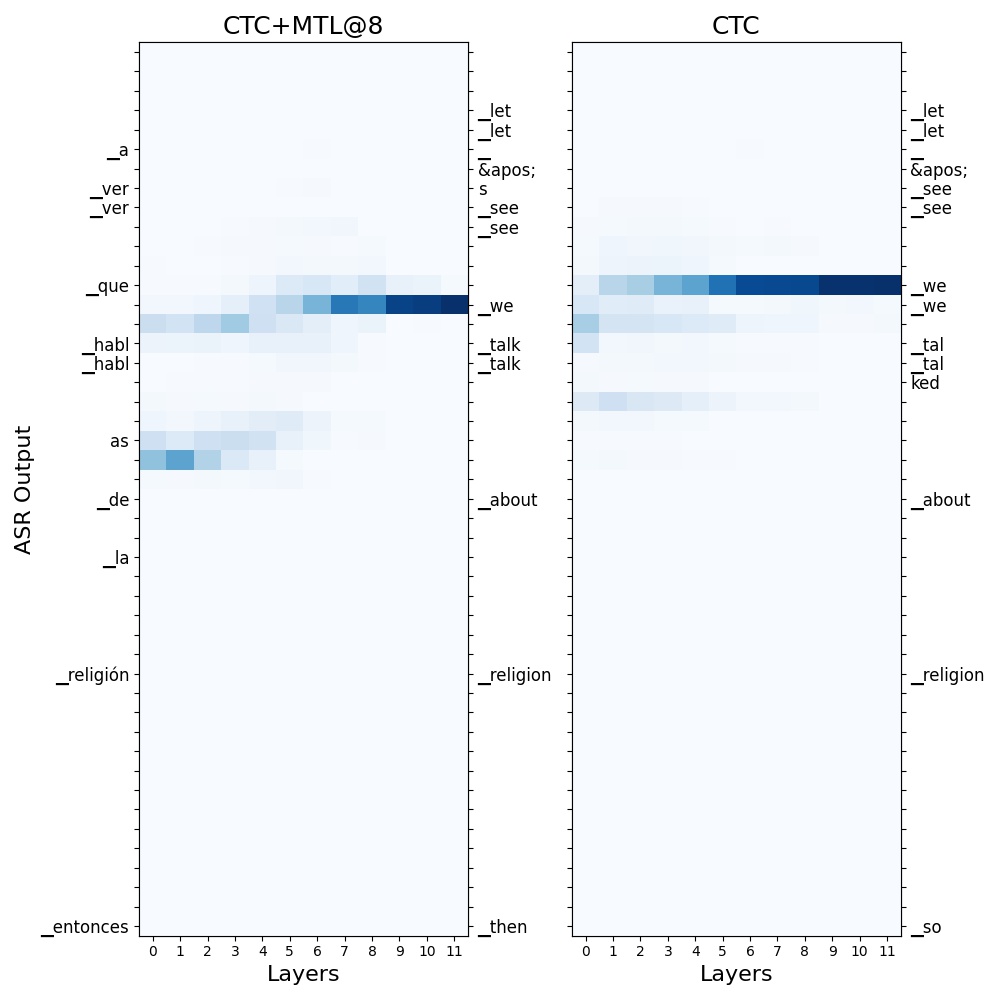}
    \caption{The re-ordering matrix of the token "\texttt{\_we}".}
    \label{fig:reorder_we}
\end{figure}

\begin{figure}[ht!]
    \centering
    \includegraphics[width=1.0\linewidth]{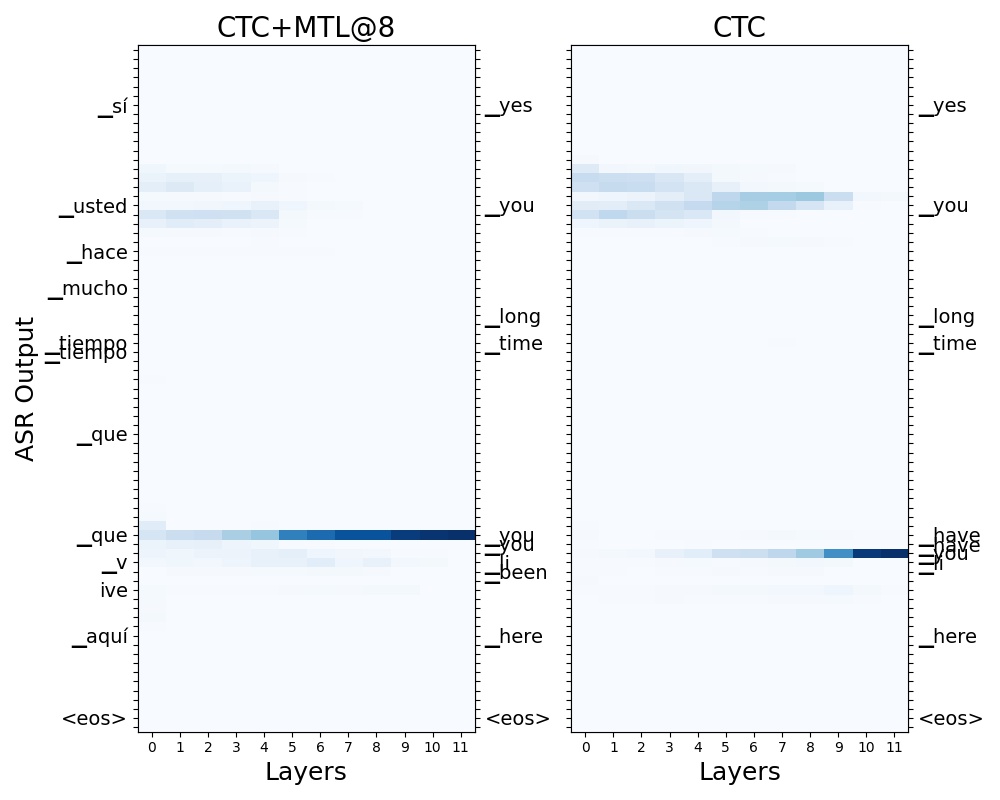}
    \caption{The re-ordering matrix of the token "\texttt{\_you}".}
    \label{fig:reorder_you}
\end{figure}

We provide some additional examples of visible reordering in our CTC-based models. In Figure~\ref{fig:reorder_we}, "\texttt{\_we}" is heavily influenced by the position of audio signal "amos", even though the ASR output is incorrectly predicted as "as". In Figure~\ref{fig:reorder_you}, "\texttt{\_you}" also influenced by audio signal "usted". It is interesting to observe that in some cases the pure CTC-based model appears more capable of reordering, while in others it does not.

\section{Higher Reordering Difficulty}
\begin{figure}[t!]
    \centering
    \includegraphics[width=\linewidth]{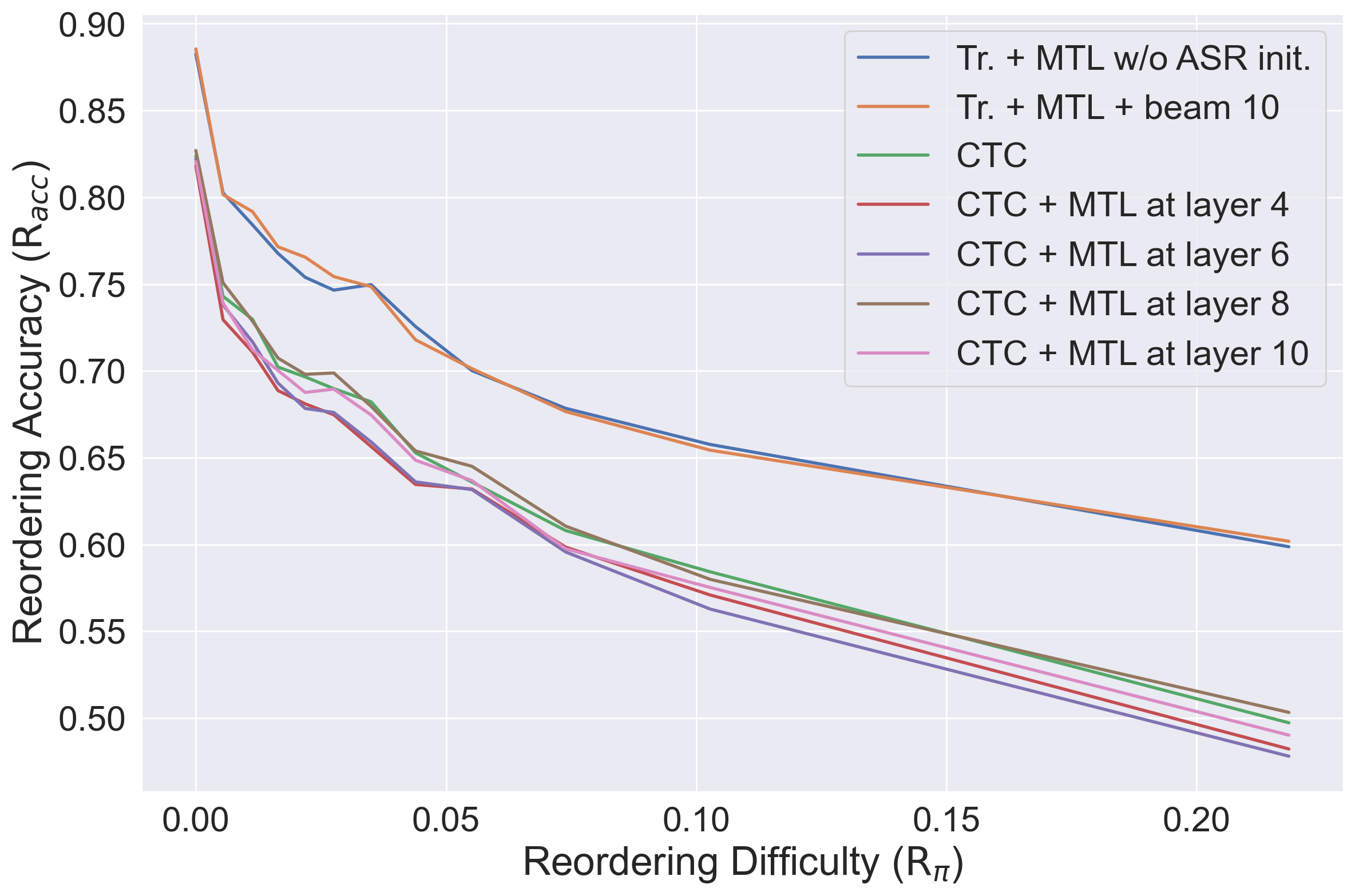}
    \caption{The reordering accuracy ($\mathrm{R}_{acc}$) curve under different reordering difficulties ($\mathrm{R}_\pi$).}
    \label{fig:acc_vs_difficulty_curve}
    \vspace{-5pt}
\end{figure}

We address instances of higher difficulty by analyzing Figure~\ref{fig:acc_vs_difficulty_curve}. In the figure, the horizontal axis corresponds to the reordering difficulty, and the vertical axis corresponds to the reordering correctness. In this figure, there is a very consistent decrease in reordering correctness when reordering difficulty increases, and the rate of decrease is very similar between NAR and AR models. This observation reveals that when evaluated on distant language pairs, that the reordering difficulty is large, the gap between NAR and AR will probably remain roughly the same. We will conduct experiments on different language pairs to verify the above claim in future work.

\end{document}